%% file: acl_latex.tex
\documentclass[11pt]{article}

\usepackage[final]{acl}

\usepackage{times}
\usepackage{latexsym}
\usepackage{booktabs}
\usepackage{makecell}
\usepackage[T1]{fontenc}

\usepackage[utf8]{inputenc}

\usepackage{pifont}   
\usepackage{amssymb}  
\definecolor{correctgreen}{RGB}{34,139,34}
\definecolor{driftred}{RGB}{200,40,40}

\usepackage{soul}                                                                           
\newcommand{\hlred}[1]{\colorbox{red!15}{#1}}
\newcommand{\hlblue}[1]{\colorbox{blue!15}{#1}} 
\usepackage{microtype}
\usepackage{inconsolata}
\usepackage{graphicx}
\usepackage{amsmath}
\usepackage{amssymb}
\usepackage{booktabs}
\usepackage{multirow}
\usepackage{xcolor}
\usepackage{algorithm}
\usepackage{algorithmic}
\usepackage{bbm}
\usepackage{hyperref}
\usepackage{url}
\usepackage{booktabs}
\usepackage{amsfonts}
\usepackage{amsmath}
\usepackage{amssymb}
\usepackage{amsthm}
\usepackage{nicefrac}
\usepackage{microtype}
\usepackage{xcolor}
\usepackage{graphicx}
\usepackage{algorithm}
\usepackage{algorithmic}
\usepackage{multirow}
\usepackage{subcaption}
\usepackage{amsthm}
\usepackage{enumitem}
\usepackage{svg}
\usepackage{caption}
\usepackage{comment}
\usepackage{tabularx}
  \usepackage{dblfloatfix}   

\theoremstyle{definition}
\newtheorem{definition}{Definition}[section]

\definecolor{baseblue}{RGB}{52, 152, 219}    
\definecolor{hybridblue}{RGB}{27, 79, 114}

 \newcommand{\ts}[1]{{\,\scriptsize\textcolor{gray}{#1}}}

\title{Know When to Stop: Segment-Level Credit Assignment for Reducing Overthinking}

\author{
  \textbf{Chia-Hsuan Lee, Sihui Dai, Mingyang Zhou, Isha Slavin, Hsuan Su} \\
  \textbf{Shi-Xiong Zhang, Sambit Sahu, William Campbell} \\
  Capital One
}

\begin{document}
\maketitle
\begin{abstract}                                                                        Reasoning language models frequently overthink: generating extended chains of behaviors such as hedging, approach abandonment, and self contradiction that consume tokens without improving answers. We show that these behaviors are not merely a consequence of length; even when controlling for response length, incorrect traces exhibit higher rates of unproductive self-reflection than correct ones.  Addressing this requires identifying where self-reflection helps vs hurts, but obtaining these step-level annotations is costly.  We observe that intermediate answer commitments within reasoning traces can provide a cheap proxy: by comparing each final answer candidate in the trace to the ground truth, we can determine whether subsequent reflection is productive without any additional supervision.  Building on this insight, we propose DASH (Drift Aware advantage SHaping), which assigns segment-level credit based on whether each reasoning segment leads toward or away from correctness.  On competition-level math benchmarks, DASH achieves the highest accuracy where overthinking is prevalent (Average Accuracy: 59.45\% vs. 58.1\% Dr.GRPO vs. 56.95\% GRPO) while reducing overthinking behaviors and achieving more productive self-correction than baselines.
                                         
  \end{abstract}

\input{sections/introduction}
\input{sections/overthinking_signals}
\input{sections/drfit_aware_reward}

\input{sections/experiment}

\input{sections/analysis}

\input{sections/related}
\input{sections/conclusion}


\section*{Limitations}
Our work has several limitations. First, majority of experiments are conducted on a 4B model; while our analysis suggests drift patterns are scale-invariant, training dynamics may differ at larger scales. Second, our method requires extractable intermediate answers for drift detection, which limits applicability to domains with verifiable checkpoints (mathematics, code with test cases). Open-ended reasoning tasks without clear answer markers would require alternative drift indicators. Third, the slight accuracy decrease on some easy benchmarks for Nemotron (MATH-500: $-1.7$pp vs.\ base) represents a real trade-off that practitioners must weigh against gains on hard problems. Finally, our evaluation is limited to mathematical reasoning; generalization to other reasoning domains (logical, scientific, commonsense) remains to be validated. 

Future work could explore difficulty-adaptive methods that modulate drift penalty strength based on problem difficulty or model uncertainty.

As with any RL fine-tuning method, DASH inherits the safety properties and failure modes of the base model; we do not introduce safety mitigations beyond those of the upstream models.

\section*{Ethics Considerations}
This work studies segment-level credit assignment for reducing overthinking in mathematical reasoning models. We use publicly available mathematical datasets and open-weight models, and no human subjects or personal data were involved. All training signals are derived from model-generated reasoning traces and verifiable ground-truth answers, without human annotation or external learned judges. We do not foresee direct negative societal impacts from the methodological contributions of this paper, as the work focuses on scientifically understanding and improving reasoning behavior on mathematical benchmarks. The resulting models may inherit limitations from the underlying models and datasets. We also acknowledge that reinforcement-learning experiments require substantial computational resources, although environmental impact was not directly measured. 

\bibliography{custom}

\appendix
\input{sections/appendix_exp}

\end{document}

%% file: sections/introduction.tex
\section{Introduction}
\label{sec:intro}

\begin{figure*}
    \centering
    \includegraphics[width=\linewidth]{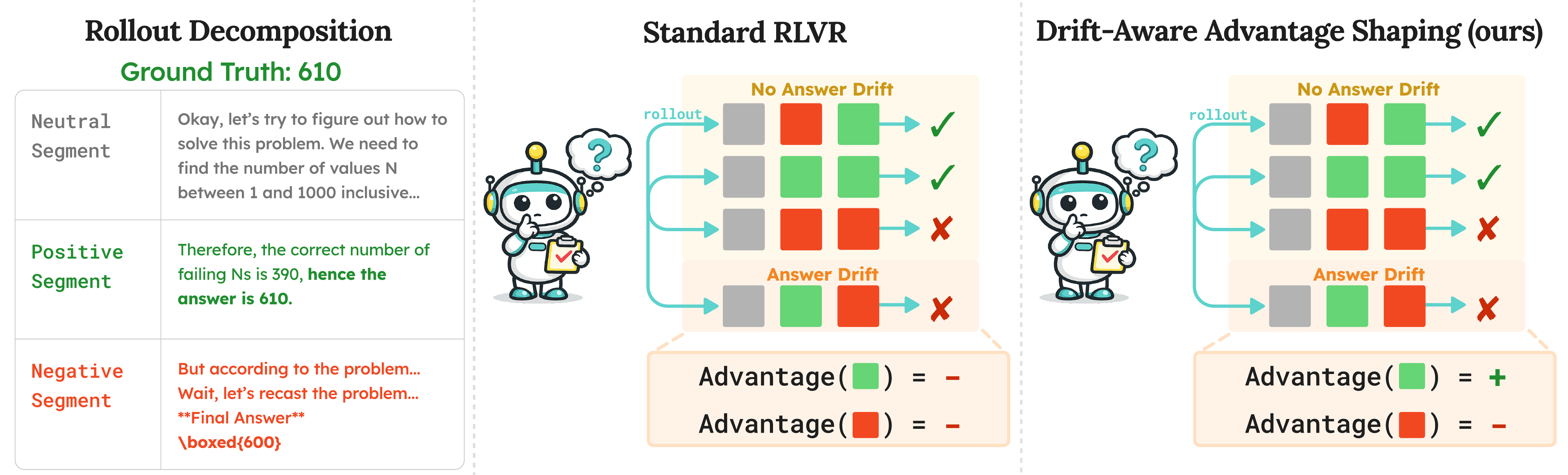}
    \caption{\textbf{Overview of DASH.} We decompose reasoning traces into segments bounded by intermediate answer checkpoints. Segments leading to correct answers (green) receive positive advantage; segments leading to incorrect answers (red) receive escalating negative advantage. Standard GRPO assigns uniform negative advantage to all tokens in an incorrect trace, discarding the structure within.}
    \label{fig:overview}
\end{figure*}

Reasoning-focused language models, such as DeepSeek-R1 \citep{deepseekai2025deepseekr1}, achieve strong performance through extended chains of thought. However, longer reasoning does not always help: models frequently exhibit overthinking behaviors such as hedging, re-verifying, or switching approaches \citep{wang2025thoughts, peng2025selfdoubt} which can lead the model to an incorrect final answer. 

This motivates a natural question: \textit{can we train models to retain productive self-reflection while surpressing unproductive self-reflection?}  A significant challenge is cost: identifying whether self-reflection is helpful at each reasoning step would require step-wise labels via a process reward model \citep{lightman2024let,wang-etal-2024-math}, LLM-as-a-judge, or manual annotation \citep{lightman2024let}.  In this work, we propose a cheaper alternative.

Our key observation is that reasoning models can commit to intermediate answers within their thinking traces--for example, writing "the answer is X" or
  boxing a result before continuing to reason. These commitments provide verifiable demonstrations of productive and unproductive self-reflection: by comparing each to the ground truth, we know whether
  subsequent reflection improved or degraded the answer, without any external supervision. When a model reaches a correct intermediate answer and
  then reflects its way to an incorrect one, we have direct evidence that this self-reflection was harmful.


Based on this intuition, we propose \textbf{DASH} (Drift-Aware advantage SHaping), which converts traces where the answer drifts from correct to incorrect intermediate examples from wasted negatives into informative training examples.  Rather than assigning a single scalar advantage to the entire rollout, DASH divides each trace into segments bounded by consecutive answer checkpoints and assigns advantages based on whether each segment moves towards or away from the correct answer.  A single drift trace simultaneously teaches the model to
  reinforce the reasoning that found the correct answer \textit{and} to suppress the overthinking that abandoned it--extracting dual training signal from
  what GRPO would treat as a flat negative example.


We complement DASH with six lightweight \textbf{linguistic overthinking signals}---repetition, hedging, abandonment, contradiction, recomputation, and length outlier---that characterize reasoning quality without requiring intermediate answer extraction. These signals serve as evaluation metrics to verify that accuracy gains correspond to genuine behavioral improvements rather than superficial shortcuts.

Experiments across four competition-level math benchmarks show:
\begin{itemize}[nosep]
    \item \textbf{Best accuracy where drift is severe.} DASH achieves 59.45\% across the challenging math evaluation suite (vs.\ 56.95\% GRPO, 58.13\% Dr.GRPO, 55.65\% base)---the benchmark with severe drift prevalence.
    \item \textbf{Self-correction over spiraling.} DASH's correct traces exhibit twice as many contradiction-then-resolution patterns as GRPO's (0.92 vs.\ 0.47/trace), while showing fewer blind approach abandonments---reasoning longer but more productively.
\end{itemize}

%% file: sections/overthinking_signals.tex
\section{Analyzing Overthinking in Reasoning Traces}
\label{sec:overthinking}
\label{sec:signals}

Prior work on reasoning efficiency has primarily characterized overthinking through response length: longer traces are treated as less efficient, and length penalties or early-stopping mechanisms are used to encourage brevity~\citep{muennighoff2025s1}.  However, length alone does not provide detail signals for what overthinking behaviors are exhibited by the model.

To address this, we first introduce a set of linguistic signals to analyze overthinking patterns. All signals are regex- or $n$-gram-based, requiring no learned components:

\begin{itemize}[nosep, leftmargin=*]
    \item \textbf{S1 (Repetition):} Maximum $n$-gram overlap between sliding windows. Captures circular reasoning loops where the model rephrases without progressing \citep{duan2026circular}.
    \item \textbf{S2 (Hedging):} Density of uncertainty markers (``wait, no,'' ``let me reconsider'') per 100 tokens. Operationalizes the self-doubt mechanism preceding negative flips \citep{zhou2026when}.
    \item \textbf{S3 (Abandonment):} Count of explicit strategy switches (``this approach is wrong,'' ``let me try another way''). The strongest individual failure predictor: 4.1--4.3$\times$ more common in incorrect traces across all models tested.
    \item \textbf{S4 (Contradiction):} Count of self-contradiction markers (``which is impossible,'' ``can't be right''). Captures unresolved inconsistencies \citep{muendler2024selfcontradictory, yang2026batched}.
    \item \textbf{S5 (Recomputation):} Numerical values derived 3+ times in computation contexts. Targets confirmatory re-checking that rarely catches errors \citep{long2026selfverification}.
    \item \textbf{S6 (Length outlier):} Per-prompt group $z$-score of response length. Adaptive: hard problems warrant long responses, but within-group outliers indicate pathology.
\end{itemize}

Using these signals, we analyze reasoning traces generated by Llama-3.1-Nemotron-Nano-4B-v1.1 \citep{nvidia2025llamanemotron} on AIME 2025.  We bucket generated responses by length and analyze the correlation between correctness, length, and overthinking behaviors.  We present results for abandonment and self-contradiction dimensions in Figure \ref{fig:overthinking-signals} and a full analysis in Appendix \ref{app:full_overthinking}.

\begin{figure*}
    \centering
    \includegraphics[width=\linewidth]{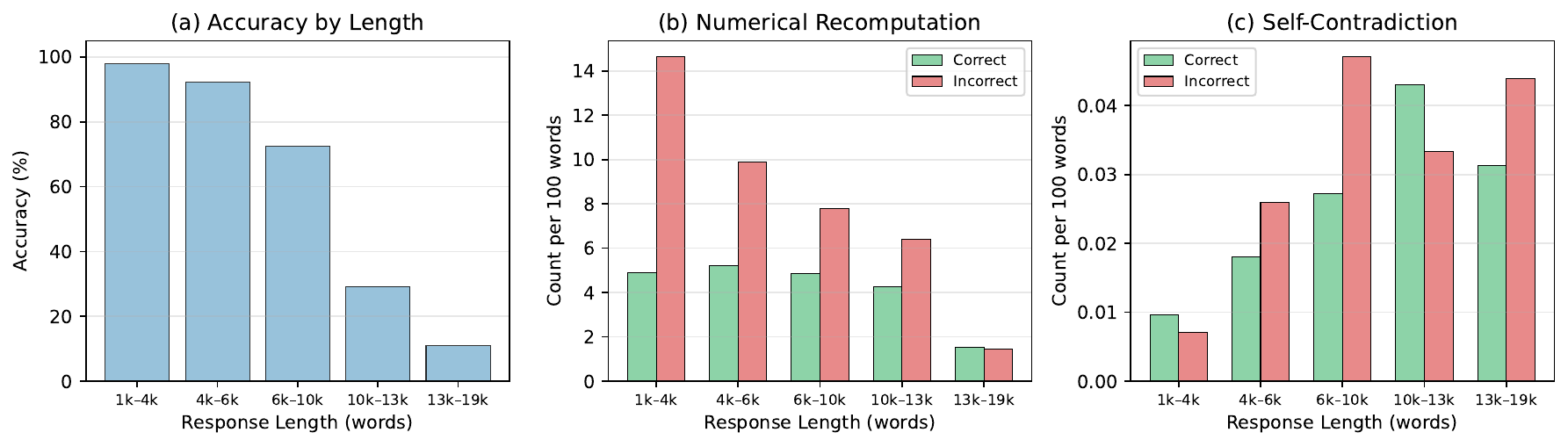}
    \caption{Overthinking signals in Nemotron-4B reasoning traces on AIME 2024 (960 traces, 32 per problem). Traces are grouped into quintiles by word count. (a) Accuracy drops sharply with response length. (b–c) Numerical recomputation and self-contradiction density (per 100 words), split by correctness within each length bucket. Even controlling for length, incorrect traces exhibit higher rates of unproductive self-reflection than correct traces, indicating these linguistic signals carry information beyond response length alone.}
    \label{fig:overthinking-signals}
\end{figure*}

Figure~\ref{fig:overthinking-signals}(a) shows that accuracy decreases monotonically with response length, confirming that the model's additional computation in longer traces is largely unproductive. Panels (b) and (c) reveal a more nuanced finding: even controlling for response length,   incorrect traces exhibit higher rates of numerical recomputation and self-contradiction than correct traces within the same length bucket. The gap
  is most pronounced for numerical recomputation, where incorrect traces show consistently elevated density across all buckets, while
  self-contradiction shows a more moderate trend.

  These patterns suggest that much of the model's unproductive computation involves cycles of self-reflection that actively steer reasoning in the
  wrong direction—repeating derivations or arriving at contradictions that undermine earlier progress. This motivates our central
  hypothesis: if we can discourage self-reflective behavior that ultimately leads to incorrect answers while preserving reflection that aids error
  correction, we may improve both reasoning efficiency and accuracy without simply truncating generation length.


%% file: sections/drfit_aware_reward.tex

\section{Drift-Aware Advantage Shaping}
\label{sec:method}
One challenge is that it is difficult to identify where self-reflective behavior is needed for correcting errors in reasoning steps and where it hurts performance.  In this section, we consider a cheap proxy: rather than obtaining annotations across each step containing self-reflection and whether it is helpful or not, we instead extract signals from self-reflection that occurs \textit{after} arriving at a candidate final answer, which can be verified with the ground truth as a measure of whether self-reflection was helpful or not.

   



\subsection{Preliminaries: GRPO}

In Group Relative Policy Optimization \citep{shao2024deepseekmath}, for each prompt $x$, the model generates $n$ rollouts $\{y_1, \ldots, y_n\}$. Each rollout receives a reward $r_i$, and advantages are computed by group normalization:
\begin{equation}
    A_i = \frac{r_i - \text{mean}(\{r_j\}_{j=1}^n)}{\text{std}(\{r_j\}_{j=1}^n)}
\end{equation}

This scalar $A_i$ is broadcast identically to every token position in rollout $y_i$, yielding per-token advantages $a_t = A_i$ for all $t \in \{1, \ldots, |y_i|\}$. The policy gradient loss is:
\begin{equation}
    \mathcal{L} = -\frac{1}{|y_i|} \sum_{t=1}^{|y_i|} \min\left(\rho_t A_i,\ \text{clip}(\rho_t, 1\pm\epsilon) A_i\right)
\end{equation}
where $\rho_t = \pi_\theta(y_t | x, y_{<t}) / \pi_{\text{ref}}(y_t | x, y_{<t})$.

\subsection{Identifying Answer Drift}
Our proposed algorithm is centered around the key idea of detecting portions of self-reflection where the final answer is incorrect even though the trace had reached the correct solution at some point.  Formally, we define answer drift as:
\begin{definition}[Answer Drift]
\label{def:drift}
Let $y$ be a reasoning trace produced for a question with ground-truth answer $a^*$. Suppose we extract from $y$ an ordered sequence of intermediate answer candidates $(\hat{a}_1, \ldots, \hat{a}_K)$, representing successive points at which the model commits to an answer before potentially reconsidering. We say $y$ exhibits \emph{answer drift} if $\hat{a}_K \ne a^*$ and there exists some $i \in \{1, \ldots, K-1\}$ such that $\hat{a}_i = a^*$.
\end{definition}

In order to identify drift occurrence, we extract intermediate answer commitments within the reasoning trace by matching patterns including $\backslash$\texttt{boxed\{...\}}, ``the answer is $X$'', 
and natural-language answer commitments.  Instances where drift occurs are prime examples of where self-reflection harms performance that we aim to penalize.

\subsection{Segment-Based Advantage Shaping}
\label{sec:segment}
One important consideration is that we would like to maintain positive self-correction even within traces with drift.  For example, a trace can oscillate between a correct answer candidate and an incorrect answer candidate, and we would like to encourage behavior where the model's self reflection changes to the correct answer while penalizing shifts from correct to incorrect.  To handle this, rather than broadcasting a single advantage to all tokens, we divide each rollout into \emph{segments} bounded by consecutive answer checkpoints ($\hat{a}_{j-1}, \hat{a}_{j}$] and assign segment-specific advantages.

\paragraph{Segment construction.} Given checkpoints at token positions $p_1 < p_2 < \ldots < p_K$ within the response, we define segments:
\begin{itemize}[nosep, leftmargin=*]
    \item \textbf{Neutral segment} $S_0$: tokens before the first checkpoint ($t < p_1$)
    \item \textbf{Positive segment} $S_j^+$: tokens in segment $(\hat{a}_{j-1}, \hat{a}_{j}]$ where checkpoint $\hat{a}_{j}$ is correct
    \item \textbf{Negative segment} $S_j^-$: tokens in segment $(\hat{a}_{j-1}, \hat{a}_{j}]$ where checkpoint $\hat{a}_{j}$ is incorrect
\end{itemize}

\paragraph{Advantage assignment.} For each token at position $t$:
\begin{equation}
    a_t = \begin{cases}
        +|A_i| \cdot \alpha_+ \cdot d_j & \text{if } t \in S_j^+ \\
        -|A_i| \cdot \alpha_- \cdot w(t) & \text{if } t \in S_j^- \\
        A_i \cdot \alpha_n & \text{if } t \in S_0 \text{ (conditional)}
    \end{cases}
    \label{eq:segment_advantages}
\end{equation}

where $\alpha_+, \alpha_-$ are positive and negative scale factors, $\alpha_n$ is the neutral scale, $d_j$ is a diminishing-return weight for repeated confirmation, and $w(t)$ is a length penalty weight, both defined below.

\paragraph{Diminishing returns for repeated confirmation.} Drift-prone models frequently re-derive and re-confirm the same correct answer several times before drifting away from it; rewarding every confirmation equally would itself reinforce overthinking. We therefore attenuate consecutive positive segments geometrically:
\begin{equation}
    d_j = \max\left(\gamma^{\,k_j},\ \gamma_{\min}\right)
\end{equation}
where $k_j$ counts the consecutive positive segments immediately preceding $S_j^+$ and resets to zero whenever a negative segment intervenes. The first arrival at a correct answer thus receives full credit ($d_j = 1$), redundant re-confirmations receive geometrically decaying credit, and the floor $\gamma_{\min}$ preserves a weak signal even for highly redundant segments.

\paragraph{Length penalty within negative segments.} To encode ``the longer you continue past a correct answer, the worse it gets,'' tokens in negative segments receive an escalating penalty:
\begin{equation}
    w(t) = 1 + \alpha \cdot \frac{t - t_{\text{start}}^{(k)}}{t_{\text{end}}^{(k)} - t_{\text{start}}^{(k)}}
\end{equation}
where $\alpha$ controls the ramp rate and the penalty is capped at $w_{\max}$ to prevent gradient explosion.

\paragraph{Conditional treatment of pre-answer reasoning.} Tokens before the first answer checkpoint ($S_0$) present a design choice: in a drift trace, this reasoning successfully produced a correct answer---the failure was in not stopping afterward---so penalizing it alongside the drift suffix would discourage valid reasoning. We therefore condition $\alpha_n$ on the trace outcome: correct traces receive the standard advantage, drift traces receive a weak positive signal, and pure-incorrect traces receive no gradient on $S_0$.


\subsection{Reward Shaping for Drift Traces}
\label{sec:reward_shaping}

Before advantage computation, drift traces receive a \emph{shaped reward} that reflects their partial correctness:
\begin{equation}
    r_{\text{drift}} = r_{\text{incorrect}} + \delta \cdot \left(1 - \frac{L_{\text{post-drift}}}{L_{\text{total}}}\right)
\end{equation}
where $\delta$ is the drift partial credit and the shaped reward naturally decreases as the post-drift portion of the trace grows. This ranks drift traces strictly above pure-incorrect traces, reflecting that they demonstrated partial capability, while keeping them below correct traces.

\subsection{Generality Across GRPO Variants}
\label{sec:composition}
DASH operates purely on per-token advantages and therefore composes with any GRPO-family optimizer without algorithmic modification. When paired with DR-GRPO \citep{liu2025understanding}, only two hyperparameters change, both following directly from DR-GRPO's token-mean loss aggregation: because token-mean already supplies an implicit length signal---shorter correct traces receive proportionally stronger per-token gradients---we disable the explicit length penalty ($w(t) = 1$) to avoid double-counting, and reduce $\alpha_+$ from $1.0$ to $0.5$ so that the positive segment credit does not compound with this implicit signal. All other components transfer unchanged.

%% file: sections/experiment.tex

\section{Experimental Results}
\subsection{Experimental Setup}
\label{sec:setup}
\subsubsection{Model and Data}
We primarily experiment with \textbf{Llama-3.1-Nemotron-Nano-4B-v1.1} \citep{nvidia2025llamanemotron}. To test generalization across model families, we additionally train GRPO and GRPO+DASH on \textbf{Phi-4-reasoning-plus}
  \citep{phi4reasoning} and \textbf{OLMO-3-think} \citep{olmo2025olmo} (\S\ref{sec:generalization}). Our training data consists of 16.5K mathematical reasoning problems sampled from
   OpenR1-Math-220K~\citep{openr1}, sourced from NuminaMath~1.5~\citep{numina2024} with reasoning traces from DeepSeek-R1~\citep{deepseekai2025deepseekr1} verified by
  Math-Verify. 

\subsubsection{Evaluation}
We evaluate on four competition-level mathematical reasoning benchmarks: \textbf{OlympiadBench} \citep{he-etal-2024-olympiadbench}, \textbf{MinervaMath} \citep{lewkowycz2022solving}, \textbf{AIME 2024} \citep{aime24}, and \textbf{AIME 2025} \citep{aime25}. For AIME24, and AIME25, we report avg@32 (average accuracy over 32 sampled solutions per problem). For OlympiadBench and MinervaMath, we report pass@1.

\subsubsection{Training Configuration}
\label{sec:training_config}
All runs use GRPO with group size $n=16$ trained on 4 nodes of 8 H100 GPUs. For our drift-aware (DASH) runs, we set positive and negative advantage scales to $\alpha_+ = 1.0, \alpha_- = 1.0$, use a length penalty with $\alpha = 3.0$ and $w_{\max} = 3.0$, and apply a \texttt{conditional} neutral mode with $\alpha_n = 0.1$. Full training hyperparameters are provided in Appendix~\ref{app:training_details}.

\subsubsection{Baselines}
Along with the base model and standard GRPO, we compare against two baselines designed to address lengthy reasoning traces:

\paragraph{DR-GRPO \citep{liu2025understanding}.} DR-GRPO is a debiased variant of GRPO that addresses an implicit length bias in the optimization objective. It makes two modifications: (1) replacing per-response length normalization in the policy loss with a constant scaling factor, which ensures that longer and shorter responses receive equal per-token gradient weight, and (2) computing advantages as $\hat{A}_i = r_i - \text{mean}(\{r_j\}_{j=1}^n)$ without dividing by $\text{std}(\{r_j\}_{j=1}^n)$, which removes a question-level difficulty bias. These changes have been shown to improve token efficiency and reduce the length of incorrect responses. This baseline tests whether an explicit credit assignment over the reasoning trajectory is more effective at reducing drift and overthinking behavior compared to using a debiased optimizer.

\paragraph{GRPO + Brevity Bonus.} We introduce a simple reward shaping baseline that explicitly encourages shorter correct traces. For each prompt group, we identify the length of the shortest correct response $l_{\min}$. Each correct trace of length $l_i$ then receives an additive per-token bonus: $b_i = \frac{\beta \cdot l_{\min}}{l_i^2}$, where $\beta$ is a scale hyperparameter. The shortest correct trace receives the maximum total bonus of $\beta$, while longer correct traces receive proportionally less (scaling as $l_{\min}/l_i$). Incorrect traces receive no bonus. This baseline tests whether a simple length pressure on correct traces--without any fine-grained credit assignment over the reasoning trajectory--suffices to reduce drift and overthinking behavior.  In our experiments, we use $\beta = 0.2$
\subsection{Results}
\label{sec:results}

\subsubsection{Main Results}
Table~\ref{tab:main_results} presents accuracy across benchmarks and we summarize the key observations in the following paragraphs. 

  \begin{table*}[t]
  \centering
  \small
  \setlength{\tabcolsep}{5pt}
  \renewcommand{\arraystretch}{1.2}
  \begin{tabular}{lccccc}
  \toprule
  Method & OlympiadBench & MinervaMath & AIME24 & AIME25 & Average \\
  \midrule
  \multicolumn{6}{l}{\textit{Off-the-Shelf}} \\
  Nemotron-4B          & 60.9 & 55.1 & 60.5 & 46.1 & 55.65 \\
  \midrule
  \multicolumn{6}{l}{\textit{RL Baselines}} \\
  GRPO                 & 67.3 & 53.3 & 61.8 & 45.4 & 56.95 \\
  DR-GRPO              & 66.7 & 53.3 & 62.2 & \textbf{50.3} & 58.13 \\
  GRPO + Brevity Bonus & 66.5 & 56.6 & 58.5 & 45.9 & 56.88 \\
  \midrule
  \multicolumn{6}{l}{\textit{DASH (Ours)}} \\
  GRPO + DASH          & \textbf{68.0}\ts{+0.7} & 52.6\ts{-0.7} & 62.1\ts{+0.3} & 47.2\ts{+1.8} & 57.48\ts{+0.53} \\
  DR-GRPO + DASH       & 67.6\ts{+0.9} & \textbf{57.4}\ts{+4.1} & \textbf{65.3}\ts{+3.1} & 47.5\ts{-2.8} &
  \textbf{59.45}\ts{+1.32} \\
  \bottomrule
  \end{tabular}
  \caption{Main results. Subscripts show improvement over the corresponding base method. Best per column in
  \textbf{bold}.}
  \label{tab:main_results}
  \end{table*}

  \paragraph{(1) Strongest performance on the hard math benchmark.} DASH achieves the highest      
 accuracy on 3 out of 4 challenging math benchmarks including OlympiadBench, MinervaMath, and AIME24, outperforming GRPO and DR-GRPO.This result suggests that segment-level credit assignment is impactful on complex math benchmarks where overthinking is severe.

  \paragraph{(2) GRPO's blind spot on AIME25.} Standard GRPO improves over the base model on OlympiadBench
  (+6.4) and AIME24 (+1.3) but \emph{degrades} on AIME25 ($-$0.7). DASH avoids this regression,
  improving AIME25 by +4.7 over base. This aligns with our drift analysis: AIME25 has the highest
  correct-to-wrong drift rate, and GRPO's uniform negative advantage on drift traces inadvertently
  penalizes the valid reasoning prefix, discouraging the strategies that found the correct intermediate
  answer.

\subsubsection{Improved Self Correction}
\begin{figure}
    \centering
    \includegraphics[width=0.9\linewidth]{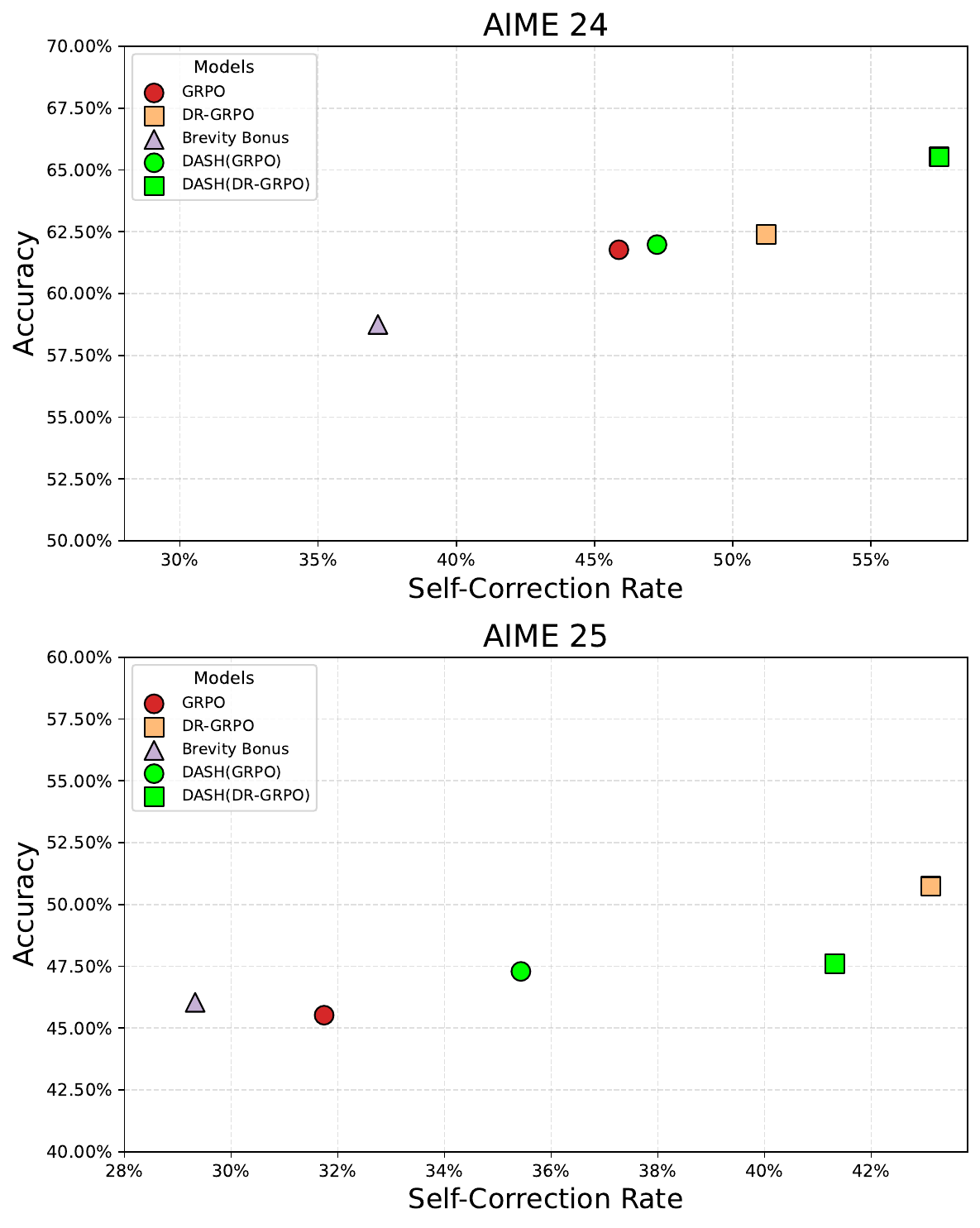}
    \caption{Self-correction rate on AIME 2024 (top) and AIME 2025 (bottom) against the benchmark accuracy. The x-axis reports the percentage of correct traces given an incorrect intermediate answer appeared before (recovery rate from incorrect answer to correct final outcome). The y-axis reports the average accuracy of different methods. Methods in the top-right region achieve both high self-correction rate and best performance of the benchmark.}
    \label{fig:drift_vs_recovery_rate}
\end{figure}
In Figure~\ref{fig:drift_vs_recovery_rate}, we plot the percentage of the positive changes in answer (Self-correction Rate: Wrong Intermediate Answer → Right Outcome) in the generated traces on AIME across DASH and baselines. We observe that conventional RL approaches like GRPO and Dr.GRPO, despite improving the base policy's accuracy on AIME tasks, still suffer from the limited capabilities to fix the incorrect intermediate answers. We observe that DASH applied on-top of GRPO and DR-GRPO, often achieves a strong self-correct capabilities which therefore leads to more accurate reasoning on challenging math-benchmarks. Brevity Bonus approach sacrifice with a significant accuracy drop on AIME tasks due to the limited self-correct capabilities.

\subsubsection{Generalizability of DASH}                                                                                                                            
  \label{sec:generalization}
    \begin{figure}
    \centering
    \includegraphics[width=0.9\linewidth]{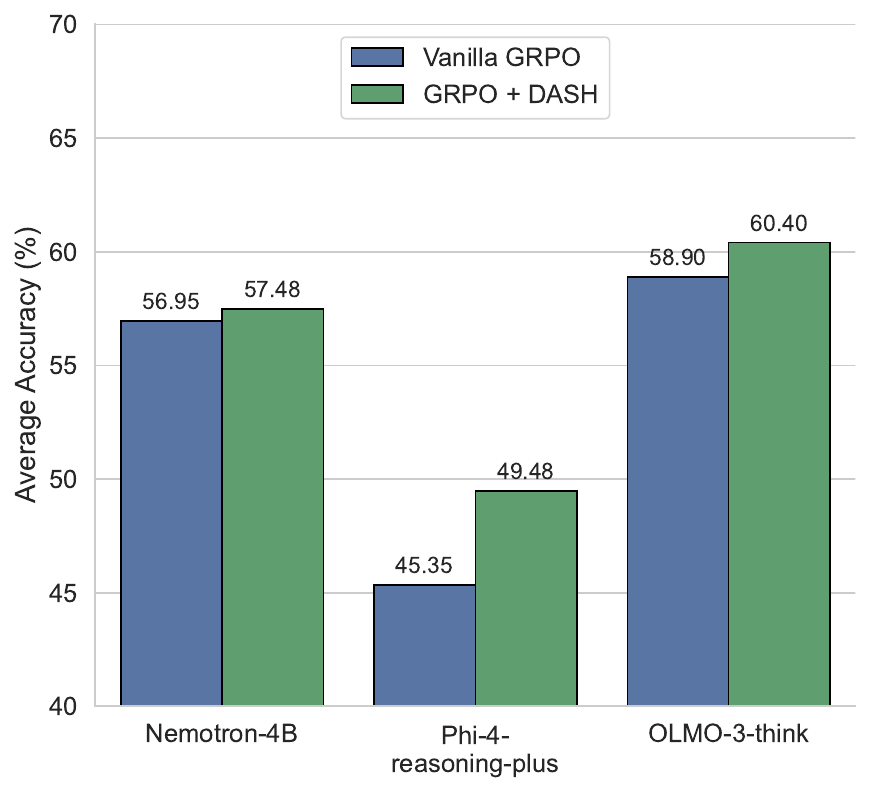}
    \caption{\textbf{DASH generalizes across model families.} Average accuracy          
    over AIME24, AIME25, Minerva-Math, and OlympiadBench for vanilla GRPO (blue) and GRPO+DASH (green) on three reasoning models from different families. GRPO+DASH improves over vanilla GRPO on all three backbones---by $+0.5$ (Nemotron-4B), $+1.5$ (OLMO-3-think), and $+4.1$ (Phi-4-reasoning-plus) points. The $y$-axis begins at 40 to make the differences legible.}
    \label{fig:generalization}
\end{figure}                                                                                                                                                                         
  To test the generalizability of DASH, we further train vanilla GRPO and               
  GRPO+DASH on two additional reasoning models from different families,
  Phi-4-reasoning-plus-14B \citep{phi4reasoning} and OLMO-3-think-SFT-7B \citep{olmo2025olmo}. We reuse the shared configuration of \S\ref{sec:training_config} and evaluate on the same four benchmarks. Since DASH draws its supervision entirely from each model's own intermediate answer commitments---with no external reward model, learned components, or per-model tuning---it applies unchanged across backbones. Figure~\ref{fig:generalization} shows that GRPO+DASH improves over vanilla GRPO on all three models: $+0.5$ on Nemotron-4B ($56.95\!\rightarrow\!57.48$), $+1.5$ on OLMO-3-think-SFT-7B ($58.90\!\rightarrow\!60.40$), and $+4.1$ on Phi-4-reasoning-plus-14B ($45.35\!\rightarrow\!49.48$). The gain is consistent across models that differ in pretraining data, scale, and tuning recipe, indicating that segment-level credit assignment from answer checkpoints is a general mechanism that improves RL training for math-reasoning. 







%% file: sections/analysis.tex
\section{Analysis}
\label{sec:analysis}






\subsection{Ablation Studies}

\paragraph{Component ablations.} Table~\ref{tab:ablation} isolates each DASH
  component by disabling one at a time while holding all else fixed. The full
  method attains the best average (57.5), and every ablation degrades it,
  confirming that the components are complementary rather than redundant.
  \emph{Reward shaping} is the single most important ingredient: setting the drift
  partial-credit $\delta$ to zero---which collapses the outcome signal to a
  GRPO-style binary reward---drops the average to 54.6 ($-2.9$), the largest
  decline and the only configuration that fails to improve over training.
  \emph{Length penalty} is next most impactful ($-2.2$, to 55.3), consistent with
  its role in curbing the length-driven overthinking that DASH targets. The
  remaining two components have small aggregate effect on accuracy: removing
  \emph{conditional-neutral} handling costs $0.8$ points (56.7) and removing the
  \emph{diminishing decay} on consecutive positive segments costs $0.7$ (56.8).
  Their benchmark-level effect is mixed---each even surpasses the full model on a
  single split (AIME25, 50.0 and Minerva-Math, 54.8, respectively)---indicating
  that these refinements primarily sharpen reasoning efficiency and stability
  (Figure~\ref{fig:radar}) rather than raw accuracy. In aggregate, reward shaping
  and the length penalty drive the bulk of DASH's gains.

\begin{table*}[t]
  \centering
  \small
  \begin{tabular}{lccccc}
  \toprule
  Variant & OlympiadBench & Minerva-Math & AIME24 & AIME25 & Average \\
  \midrule
  \textbf{DASH (full)}          & \textbf{68.0} & 52.6          & \textbf{62.1} & 47.2          & \textbf{57.5} \\
  \quad w/o conditional neutral & 63.3          & 52.2          & 61.3          & \textbf{50.0} & 56.7 \\
  \quad w/o diminishing returns   & 64.9          & \textbf{54.8} & 61.5          & 46.0          & 56.8 \\
  \quad w/o length penalty on negative     & 61.6          & 51.8          & 61.3          & 46.4          & 55.3 \\
  \quad w/o reward shaping & 60.6        & 50.7          & 60.7          & 46.2          & 54.6 \\
  \bottomrule
  \end{tabular}
  \caption{Ablation study on DASH components. Best per column in \textbf{bold}.
  Average is computed over OlympiadBench, Minerva-Math, AIME24, and AIME25.}
  \label{tab:ablation}
  \end{table*}

\subsection{Overthinking Signal Analysis}
  \label{sec:overthinking_analysis}

  \begin{figure}[]
      \centering
      \includegraphics[width=\columnwidth]{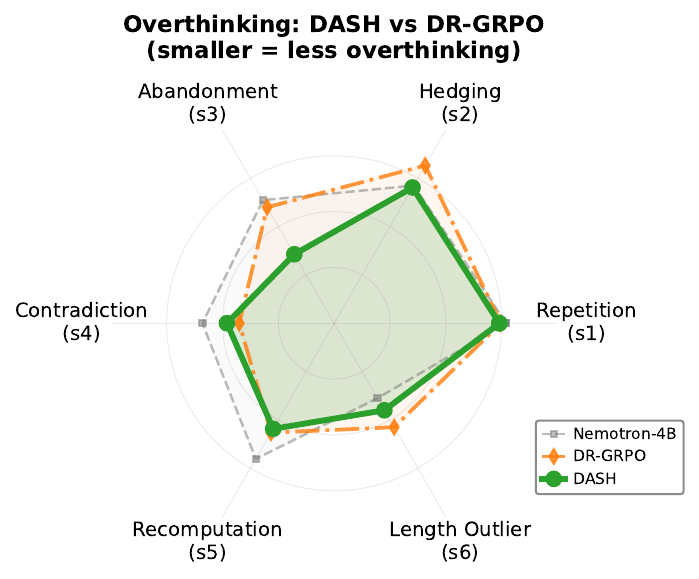}
      \caption{\textbf{Overthinking signal profile (averaged across AIME24, AIME25, OlympiadBench, and Minerva
  Math).} Each axis represents one of six linguistic signals, normalized to $[0,1]$. Smaller area = less
  overthinking. DASH reduces hedging by 14\%, abandonment by 41\%, and length outliers by 17\% relative to DR-GRPO,
  while achieving the highest accuracy (59.4\% avg@32). The sole elevated signal---contradiction (s4)---reflects
  productive self-monitoring (\S\ref{sec:contradiction_appendix}).}
      \label{fig:radar}
  \end{figure}

  Figure~\ref{fig:radar} compares the overthinking signal profile of DASH against the OTS model and DR-GRPO
  baseline, averaged across four benchmarks. DASH achieves the tightest profile: lowest abandonment (s3: 1.07 vs.\
  1.80 DR-GRPO, $-$41\%), lowest hedging density (s2: 0.53 vs.\ 0.61 DR-GRPO, $-$14\%), and lowest length-outlier
  rate (s6: 0.004 vs.\ 0.004 DR-GRPO, $-$17\%)---while maintaining nearly identical response lengths (+0.03\%) and
  achieving the highest accuracy (59.4\% vs.\ 58.1\% DR-GRPO avg@32).

  \paragraph{Contradiction as self-monitoring.} The one axis where DASH exceeds DR-GRPO is contradiction (s4: 0.96
  vs.\ 0.85, +13\%). Rather than indicating overthinking, this reflects a qualitative shift in error-handling
  strategy. Table~\ref{tab:contradiction} shows that on three of four benchmarks, elevated contradiction counts
  co-occur with accuracy gains---the model detects more errors in its reasoning \emph{and} successfully resolves
  them. On Minerva Math, DASH achieves \emph{both} lower contradiction counts and higher accuracy, suggesting it
  avoids unproductive contradiction loops entirely on easier problems. DR-GRPO instead responds to confusion with
  hedging and approach abandonment (the two signals DASH most strongly reduces), cycling through strategies without
  identifying what went wrong. Full discrimination statistics are in Appendix~\ref{sec:contradiction_appendix}.

  In summary: DASH's reasoning is the same length as DR-GRPO but \emph{qualitatively different}---it exhibits less
  hedging, far less abandonment, fewer length outliers, and more deliberate error-checking, consistent with a model
  that has learned to self-correct decisively rather than spiral.


%

\subsection{Qualitative Analysis}

To demonstrate how our drift-aware algorithm mitigates answer drift, we analyze two concrete case studies from the AIME 2025 evaluation set in Table~\ref{tab:drift_case_studies}. In both instances, the base model successfully uncovers the correct solution in its intermediate reasoning trace but drifts away from the answer (often numerous times), choosing alternative approaches. It ultimately exhausts the token limit without generating a final answer. In contrast, our drift-hybrid model stabilizes after a single validation phase and successfully commits to the correct solution.


%% file: sections/related.tex
\section{Related Work}
\label{sec:related}

\subsection{Overthinking and Reasoning Efficiency}
Excessive reasoning in LLMs has been documented across multiple studies. \citet{chen2024overthinking} first named the ``overthinking'' phenomenon in o1-like models, showing that models over-allocate compute to simple problems. \citet{su2025between} demonstrated a U-shaped relationship between reasoning length and accuracy, and the ``Reasoning Completion Point'' framework \citep{wei2025stopspinning} formally showed that once a model reaches its peak correctness probability, continued reasoning almost never improves the answer---it primarily re-confirms or flips to wrong. \citet{wang2025thoughts} link drift to thought-switching, and \citet{peng2025selfdoubt} trace the mechanism to self-doubt after correct answers. Rather than hoping continued reasoning self-corrects, we train the model to recognize when to stop \emph{within} a trace. We report the rest of related work in Section~\ref{app:related}.

%% file: sections/conclusion.tex
\section{Conclusion}
We presented DASH, a segment-level credit assignment method for reducing overthinking in reasoning language models. The core idea is to use the model's own answer checkpoints as free supervision: when a trace reaches a correct answer and later moves away from it, the trace reveals both a useful reasoning prefix and a harmful reflective suffix. DASH preserves this structure by rewarding segments that lead to correct checkpoints and penalizing segments that lead away from them.

Our analysis shows why this distinction matters. Overthinking is not simply a matter of response length; even among traces of comparable length, incorrect solutions contain more abandonment and unresolved contradiction. On competition math benchmarks, DASH is most effective where this failure mode is most prevalent, achieving the best AIME25 accuracy while reducing repetition, abandonment, and length outliers. The remaining contradiction signal becomes more productive: DASH uses contradictions for diagnosis and recovery rather than as a prelude to spiraling.

These results suggest that efficient reasoning training should focus less on making models universally shorter and more on teaching them when reflection has stopped being useful. Segment-level credit from answer checkpoints offers a simple way without process labels or external judges.

%% file: sections/appendix_exp.tex

\section{Signal Implementation Details}
\label{app:signals}

\paragraph{S1: Repetition.} Window size 200 tokens, stride 50. Jaccard similarity over 5-grams between non-adjacent window pairs. Score = max similarity, clipped to $[0,1]$. Threshold: $>0.4$.

\paragraph{S2: Hedging.} Case-insensitive regex: \texttt{wait}, \texttt{hmm}, \texttt{actually}, \texttt{hold on}, \texttt{let me reconsider}, \texttt{I'm confused}, \texttt{not sure}, \texttt{on second thought}. Density = count per 100 tokens.

\paragraph{S3: Abandonment.} Regex: \texttt{this approach is wrong}, \texttt{let me try another}, \texttt{let's restart}, \texttt{going back to}, \texttt{scrapping this}, \texttt{dead end}, \texttt{alternatively}. Raw count per trace.

\paragraph{S4: Contradiction.} Regex: \texttt{contradicts the previous}, \texttt{which is impossible}, \texttt{this is impossible}, \texttt{can't be right}, \texttt{that's not possible}, \texttt{inconsistent with}, \texttt{but we just showed}. Raw count per trace.

\paragraph{S5: Recomputation.} Numeric values extracted via \texttt{[-+]?\textbackslash d+(?:\textbackslash.\textbackslash d+)?}. Values appearing 3+ times within 10-token computation contexts (near $=$, $+$, $-$, $\times$) are flagged. Count of unique repeated values.

\paragraph{S6: Length outlier.} $z$-score of thinking-section length within the GRPO prompt group. Flag threshold: $z > 2.0$. Score = $\min((z - 2) / 2,\ 1)$ for $z > 2$, else $0$.

\paragraph{Composite.} Signals are normalized to $[0,1]$: $\tilde{s}_1 = \max(s_1 - 0.2, 0) / 0.8$, $\tilde{s}_2 = \min(s_2/3, 1)$, $\tilde{s}_3 = \min(s_3/3, 1)$, $\tilde{s}_4 = \min(s_4/3, 1)$, $\tilde{s}_5 = \min(s_5/5, 1)$. Composite $\omega = (\tilde{s}_1 + \tilde{s}_2 + \tilde{s}_3 + \tilde{s}_4 + \tilde{s}_5) / 5$, clamped to $[0,1]$. Overthinking flag: $\omega > 0.3$.

\section{Signal Motivation and Prior Work}
\label{app:signal_motivation}

\paragraph{S1 (Repetition).} Reasoning models generate long chains of thought but often loop at low temperatures, repeating the same text \citep{pipis2025wait}. \citet{duan2026circular} classify these as circular reasoning loops driven by self-reinforcing attention mechanisms that prevent escape from local minima.

\paragraph{S2 (Hedging) and S3 (Abandonment).} \citet{zhou2026when} examine negative flips---cases where extended reasoning changes correct answers to incorrect ones. They find that explicit reconsideration (hedging) precedes approach abandonment in over 67\% of negative flip cases, and that abandonment rates increase monotonically with token count. The ``alternatively'' marker alone is 4.1--4.3$\times$ more common in incorrect Nemotron traces across all scales tested.

\paragraph{S4 (Contradiction).} Self-contradiction is a prevalent LLM failure mode: \citet{muendler2024selfcontradictory} find contradictions in 17.7\% of all ChatGPT sentences. \citet{yang2026batched} note that extended reasoning chains specifically increase opportunities for self-contradiction and degenerate outputs.

\paragraph{S5 (Recomputation).} \citet{duan2026circular} classify ``numerical loops''---where models repeatedly derive the same constants---as a distinct loop category triggered by reasoning impasses. \citet{long2026selfverification} show through large-scale analysis that models spend a substantial fraction of reasoning on confirmatory self-verification that rarely catches errors, reducing tokens by 20.3\% when suppressed.

\paragraph{S6 (Length outlier).} Our analysis of Nemotron-4B across difficulty levels reveals a bimodal length distribution on hard problems: responses either solve in $<$2K tokens (88\% accuracy) or spiral into a 10K--25K token dead zone (0.6\% accuracy). Within-group normalization captures this pathology adaptively without penalizing legitimately long solutions.

\section{Full Overthinking Signals On Nemotron-4B}
\label{app:full_overthinking}
\begin{figure*}
    \centering
    \includegraphics[width=\linewidth]{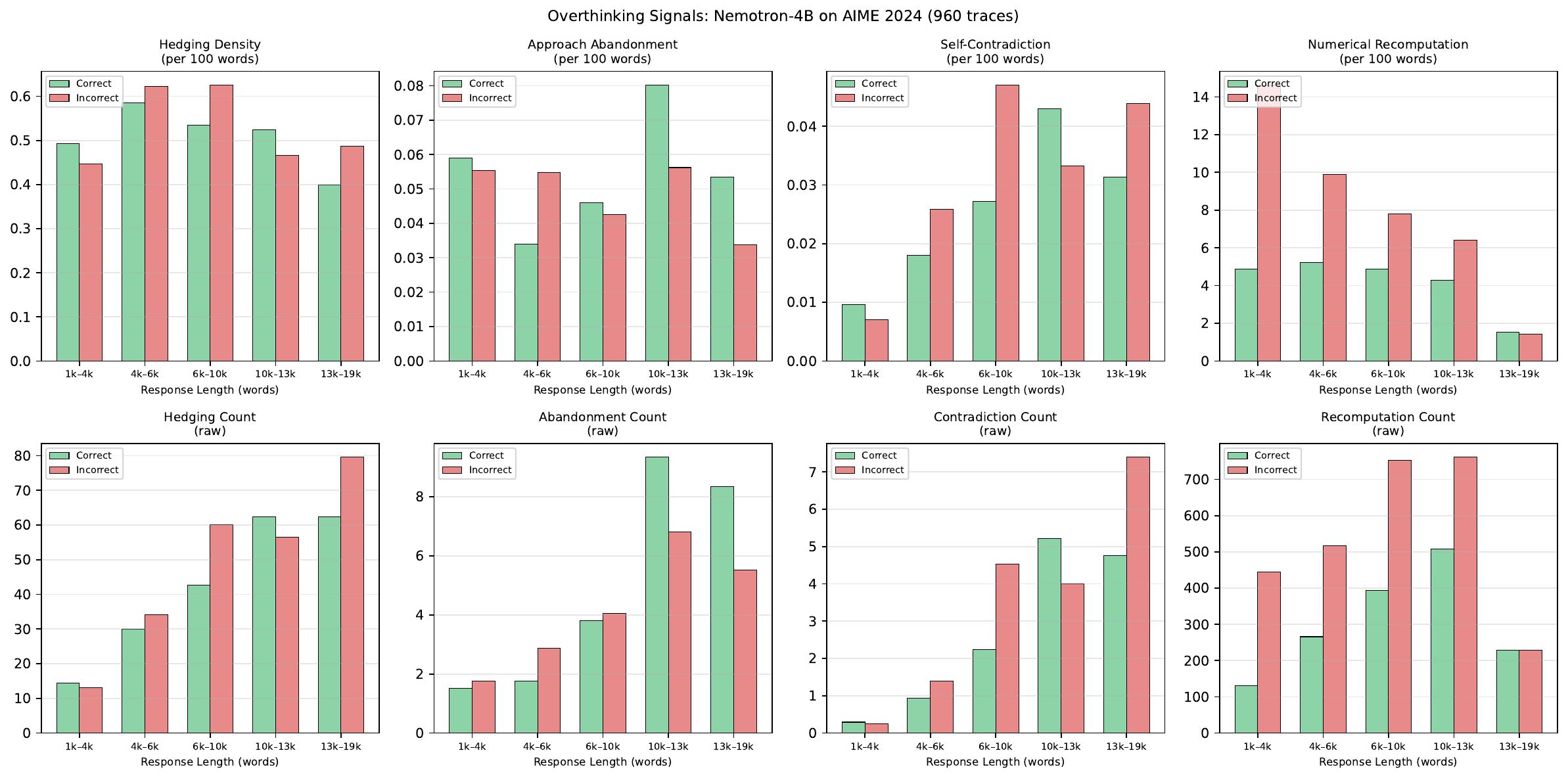}
    \caption{Full breakdown of overthinking signals in Nemotron-4B reasoning traces on AIME 2024 960 traces, 32 per problem). Traces are grouped into quintiles by word count, with bars showing the mean signal value for correct (green) and incorrect (red) traces within each bucket. Top row: density-normalized signals (counts per 100 words), which control for the trivial effect of longer traces containing more text. Bottom row: raw counts.}
    \label{fig:full_overthinking_signals}
\end{figure*}

Figure~\ref{fig:full_overthinking_signals} extends the main-text analysis by showing all four count-based linguistic signals on AIME 2024. The top row reports density-normalized counts within each length quintile, while the bottom row reports raw counts. The density panels isolate
  whether a behavior is intrinsic to failing traces or merely a side effect of their greater length: because signals are normalized per 100 words,   
  any gap between correct and incorrect bars reflects a difference in rate, not just in trace size.               
                                         
  The signal that most clearly tracks correctness is numerical recomputation. Within every length quintile except the longest, incorrect traces
  re-derive already-computed quantities at 1.5–3x the rate of correct traces — 14.6 versus 4.9 events per 100 words in the shortest bucket, and 9.9
  versus 5.2 in the next. Because this gap holds within fixed length buckets, it shows that repeated recomputation is a property of the failure mode
  itself: at a given length, traces that keep re-deriving intermediate results are markedly more likely to end incorrect. Hedging density stays near
  0.5 per 100 words regardless of correctness, and the abandonment and contradiction densities remain small in absolute terms (below 0.08 and 0.05
  per 100 words), so recomputation carries most of the density-level signal here.

  The raw-count panels show how this behavior compounds. Incorrect traces accumulate far more recomputation events in absolute terms — 444 versus 131
   in the shortest bucket, 752 versus 394 at mid-length — so failing traces are not only recomputing at a higher rate but doing so over more words,
  multiplying the effect. The picture that emerges is a model that re-opens quantitative sub-results it has already established, cycling through the
  same derivations rather than committing to them and moving forward.

  \begin{table}[t]
  \centering
  \small
  \begin{tabular}{lcccc}
  \toprule
   & \multicolumn{2}{c}{\textbf{DR-GRPO}} & \multicolumn{2}{c}{\textbf{DASH}} \\
  \cmidrule(lr){2-3} \cmidrule(lr){4-5}
  \textbf{Benchmark} & s4 & Acc. & s4 & Acc. \\
  \midrule
  AIME 2024       & 0.72 & 62.2 & 0.89 & \textbf{65.3} \\
  AIME 2025       & 0.54 & \textbf{50.3} & 0.72 & 47.5 \\
  OlympiadBench   & 0.67 & 66.7 & 0.96 & \textbf{67.6} \\
  Minerva Math    & 1.48 & 53.3 & 1.29 & \textbf{57.4} \\
  \midrule
  Average         & 0.85 & 58.1 & 0.96 & \textbf{59.4} \\
  \bottomrule
  \end{tabular}
  \caption{\textbf{Contradiction signal (s4) vs.\ accuracy.} DASH exhibits elevated contradiction counts on three
  of four benchmarks, yet achieves higher accuracy on all but AIME 2025---indicating contradictions are
  predominantly productive (detect error $\to$ resolve $\to$ correct). On Minerva Math, DASH achieves \emph{both}
  lower s4 and higher accuracy, suggesting it avoids unproductive contradiction loops entirely on easier problems.}
  \label{tab:contradiction}
  \end{table}
  
\section{Contradiction Signal: Discrimination Analysis}
  \label{sec:contradiction_appendix}

  The raw s4 (contradiction count) conflates two distinct phenomena:

  \begin{itemize}[nosep]
      \item \textbf{Productive contradiction} (predominant in DASH correct traces): detect inconsistency $\to$
  diagnose source $\to$ resolve $\to$ arrive at correct answer.
      \item \textbf{Unproductive contradiction} (predominant in baseline incorrect traces): notice error $\to$ fail
  to diagnose $\to$ silently abandon $\to$ repeat or drift.
  \end{itemize}

  Table~\ref{tab:contradiction} demonstrates that DASH's elevated s4 is predominantly productive: across four
  benchmarks, DASH increases contradiction counts by 13\% relative to DR-GRPO (0.96 vs.\ 0.85 average) while
  simultaneously improving accuracy by 1.3 percentage points (59.4\% vs.\ 58.1\%).

  The per-benchmark pattern is revealing. On the two hardest benchmarks (AIME 2024, AIME 2025), DASH increases
  contradictions substantially (+24\% and +33\% respectively). On AIME 2024, this accompanies a 3.1pp accuracy
  gain---the model detects more errors in its reasoning \emph{and} successfully resolves them. On Minerva Math (the
  easiest benchmark), DASH actually \emph{reduces} contradictions by 13\% while improving accuracy by 4.1pp,
  suggesting that on simpler problems DASH avoids entering contradiction loops altogether.

  This pattern is consistent with DASH's training objective: segment-level credit assignment rewards reasoning
  segments that lead to correct checkpoints, including segments where the model identifies and corrects an error.
  The diminishing-return decay further ensures that the model is not rewarded for repeatedly re-confirming the same
  answer, channeling self-reflection toward genuine error detection rather than redundant verification.
\subsection{Qualitative Example: Self-Correction vs.\ Spiraling}

\begin{figure*}[t]
\small
\centering
\begin{tabular}{@{}p{0.47\textwidth}|p{0.47\textwidth}@{}}
\toprule
\textbf{GRPO --- INCORRECT} \hfill \textit{61K chars} & \textbf{DASH --- CORRECT} \hfill \textit{34K chars} \\
\textit{8 abandonments $\cdot$ 0 contradictions} & \textit{1 abandonment $\cdot$ 4 contradictions} \\
\midrule
\textcolor{gray}{\scriptsize [44\%]} We need integers satisfying these. \hlred{Let's try} possible pairs. $b = 36$, $d = -36$\ldots not satisfied. \newline

\textcolor{gray}{\scriptsize [51\%]} \hlred{Let's try} $b=72$, $d=72$\ldots $-864 = -1728$. Not true. Next: $b=48$, $d=108$\ldots \newline

\textcolor{gray}{\scriptsize [61\%]} \hlred{Let's try} $q=81$, $t=13$\ldots $-136 \neq -1080$. Next: $q=27$, $t=39$\ldots $p^2 = -24$. Impossible. \newline

\textcolor{gray}{\scriptsize [75\%]} \hlred{Let's try} $b=6$: $\neq 36$. $b=4$: $\neq 36$. $b=5$: $\neq 36$. \newline

\textcolor{gray}{\scriptsize [82\%]} \hlred{Let's try} $m=0.5$\ldots negative. $m=0.8$\ldots positive. Not helpful\ldots \newline

\textcolor{gray}{\scriptsize [100\%]} \textit{[truncates mid-computation]}
&
\textcolor{gray}{\scriptsize [8\%]} Not on the line connecting centers. \hlblue{That can't be.} Let me check---$\omega_1$ internally tangent at $B$, so centers must be collinear with $B$. \newline

\textcolor{gray}{\scriptsize [23\%]} Distance $O_2$ to $B$ is 3, but $\omega_2$ has radius 15. \hlblue{That can't be.} If $\omega_1$ inside $\omega_2$, then $AO_2 = 15-6=9$\ldots \newline

\textcolor{gray}{\scriptsize [37\%]} $9+15=24$, but $AB=6$. \hlblue{This is impossible.} Must re-examine internal tangency\ldots \newline

\textcolor{gray}{\scriptsize [59\%]} $OB = OA + AB = 9+6 = 15 = r_2$. \textbf{Resolved:} $B$ is 6 units beyond $A$ from $O$. \newline

\textcolor{gray}{\scriptsize [80--100\%]} $a=2b$, $5b^2=36$, area $=288/5$. $\boxed{293}$ \\
\bottomrule
\end{tabular}
\caption{\textbf{AIME25 Problem 20} (answer = 293). \textbf{Left:} GRPO cycles through 8 blind substitutions (\hlred{red}) without diagnosing the geometric misconfiguration, truncating at 61K chars. \textbf{Right:} DASH encounters 4 contradictions (\hlblue{blue}), reasons about each, resolves the tangency configuration at 59\%, and reaches the answer in 34K chars (43\% shorter).}
\label{fig:qualitative_spiral_vs_correction}
\end{figure*}

Figure~\ref{fig:qualitative_spiral_vs_correction} contrasts the two reasoning strategies on AIME25 Problem 20 (circle geometry). GRPO never identifies the root cause---a misunderstanding of internal tangency---and instead exhaustively substitutes values across 8 different parameterizations until truncation. DASH explicitly names the geometric impossibility (``that can't be''), iteratively narrows the misunderstanding, resolves it at 59\% of the trace, and proceeds linearly to the correct computation.

\section{Training Dynamics}
\label{app:training_dynamics}

\begin{figure}[t]
    \centering
    \includegraphics[width=\linewidth]{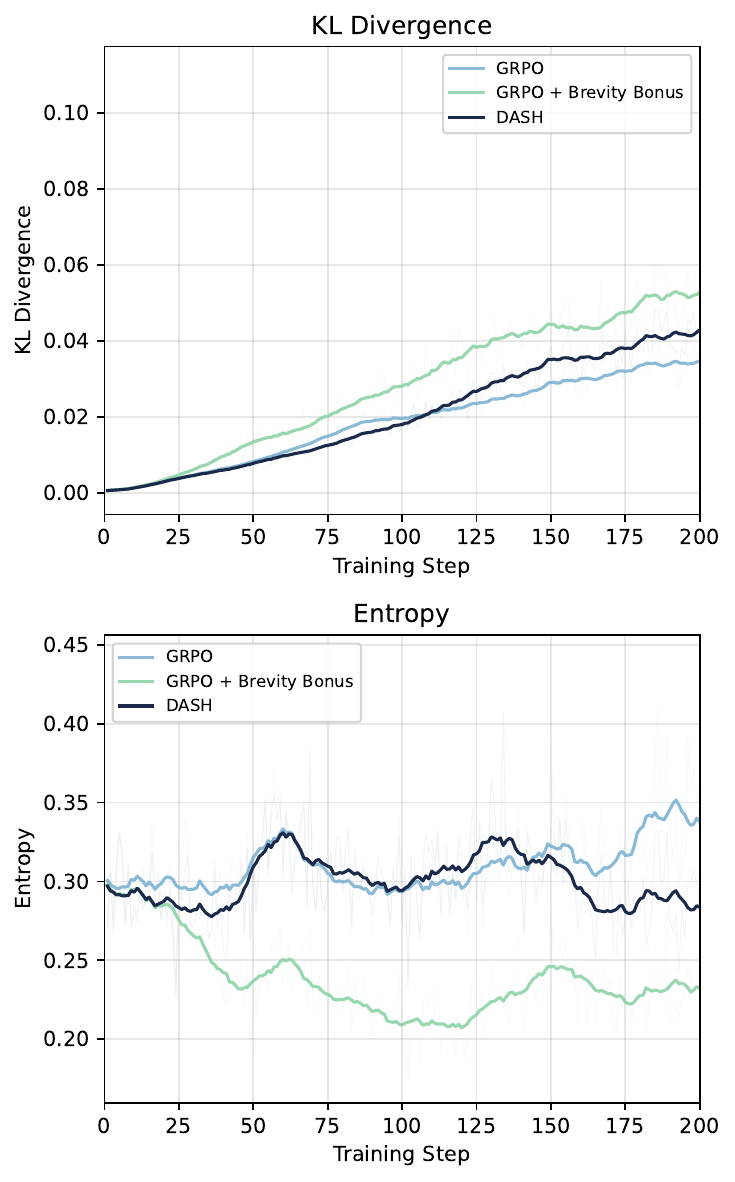}
    \caption{\textbf{KL divergence and entropy over training.} DASH maintains the lowest KL divergence, achieving length reduction through targeted credit assignment rather than aggressive policy deviation. GRPO + Brevity Bonus exhibits entropy collapse ($\to$0.21); standard GRPO shows gradually increasing entropy and KL.}
    \label{fig:kl_divergence_and_entropy}
\end{figure}

Figure~\ref{fig:kl_divergence_and_entropy} plots KL divergence and entropy over training for all methods. DASH maintains the lowest KL divergence throughout training, indicating that it achieves length reduction through targeted segment-level credit assignment rather than aggressive policy deviation, while GRPO + Brevity Bonus suffers from entropy collapse ($\to$0.21), suggesting a loss of output diversity.

\section{Training Configuration}
\label{app:training_details}

All runs share:
\begin{itemize}[nosep]
    \item GRPO with group size $n=16$, learning rate $3 \times 10^{-6}$, grad clip 1.0
    \item KL penalty: $\beta_\text{KL} = 10^{-4}$
    \item Temperature 0.9, max response length 15{,}821 tokens
    \item 4 nodes $\times$ 8 H100 GPUs, 1 epochs
\end{itemize}

\paragraph{DASH configuration.}
\begin{itemize}[nosep]
    \item Advantage estimator: \texttt{grpo\_drift\_hybrid}
    \item Positive/negative scales: $\alpha_+ = 1.0$, $\alpha_- = 1.0$
    \item Length penalty: $\alpha = 3.0$, $w_\text{max} = 3.0$
    \item Neutral mode: \texttt{conditional}, $\alpha_n = 0.1$
    \item Segment length penalty: enabled
\end{itemize}

  \section{Extended Related Work}
  \label{app:related_work}                                                                                 
                                         
  \subsection{Adaptive Reasoning Depth}
  \citet{fang2025thinkless} approach overthinking from a model-selection perspective: their Thinkless
  framework trains a model to adaptively choose between short-form and long-form reasoning via control
  tokens and a Decoupled GRPO (DeGRPO) objective, learning \emph{when} to engage in extended reasoning at
  all. Our work differs in granularity: rather than gating reasoning depth at the response level, we shape
  credit \emph{within} a single reasoning trace.

  \subsection{Additional Length Control Methods}
  \paragraph{Global vs.\ local normalization.} REINFORCE++ \citep{hu2025reinforceplusplus} replaces GRPO's
  group-level advantage normalization with global batch-level normalization, arguing that the former is a
  biased estimator that interacts poorly with length variation.

  \paragraph{Value-based methods.} VAPO \citep{yue2025vapo} addresses the length problem within a
  value-based PPO framework, introducing Decoupled GAE to handle heterogeneous sequence lengths and a
  length-adaptive discount to prevent long traces from dominating value estimates. While VAPO tackles
  length heterogeneity through the value function, our method instead operates directly on the policy
  gradient by reshaping advantages based on detected drift.

  \subsection{Token-Level Credit Assignment, Continued}
  TEMPO \citep{tran2025treecredit} builds prefix trees for nonparametric token-level credit. SPO
  \citep{guo2025segment} bridges token- and trajectory-level feedback through mid-grained, segment-level
  advantage estimation using flexible cutpoint- or tree-based partitions evaluated via Monte Carlo sampling
   without a critic model. These methods provide general-purpose credit assignment without specifically
  targeting drift; our approach is complementary in that we use the structure of intermediate answers to
  assign credit based on the model's own reasoning trajectory, requiring no additional models, sampling, or
   learned components.

  \subsection{Inference-Time Early Exit, Continued}
  Certaindex \citep{fu2024certaindex} uses answer stability for early stopping, DEER \citep{yang2025deer}
  terminates based on confidence, and ROM \citep{rom2026} monitors for real-time overthinking indicators.
  \citet{muennighoff2025s1} take a simpler approach with budget forcing, which controls test-time compute
  by forcefully terminating or extending the model's thinking process via appended tokens.

\section{Artifact Licenses}
\label{app:licenses}
 
We list below the licenses of the scientific artifacts used in this work. Our use of all artifacts is restricted to non-commercial research on language-model reasoning, which is consistent with the intended use specified by each artifact's authors.
 
\begin{itemize}[nosep]
    \item \textbf{Llama-3.1-Nemotron-Nano-4B-v1.1} \citep{nvidia2025llamanemotron}: released by NVIDIA under the NVIDIA Open Model License, with additional terms from the Llama 3.1 Community License Agreement.
    \item \textbf{veRL} \citep{sheng2025hybridflow}: released by ByteDance Seed under the Apache License 2.0.
    \item \textbf{OpenR1-Math-220K} \citep{openr1}: released by the Hugging Face Open-R1 team under the Apache License 2.0.
    \item \textbf{OlympiadBench} \citep{he-etal-2024-olympiadbench}: released by OpenBMB under the MIT License.
    \item \textbf{AMC 2023, AIME 2024, AIME 2025}: these are public mathematics competition problems originally published by the Mathematical Association of America (MAA). We use them solely as held-out evaluation benchmarks, consistent with established practice in the reasoning-LLM literature.
\end{itemize}

\section{Related Work (continued)} 
\label{app:related}
\subsection{Length Control in RL for Reasoning}
Several methods address the length problem in GRPO training.
\paragraph{Length-penalized rewards}: Dr.~GRPO \citep{liu2025understanding} corrects a per-token normalization bias that dilutes penalties for long incorrect traces; DAPO \citep{yu2025dapo} uses token-level loss aggregation; ShorterBetter \citep{yi2025shorter} introduces ``Sample Optimal Length'' as a reward signal. \paragraph{Progressive constraints}: ThinkPrune \citep{hou2025thinkprune} iteratively tightens token budgets, and L1/LCPO \citep{aggarwal2025l1} adds explicit length-controlled objectives. \paragraph{Decoupled normalization}: DRPO \citep{li2025drpo} normalizes correct and incorrect rollouts separately, preventing length effects from corrupting advantage estimates. Our method adopts decoupled normalization as a component but adds segment-level granularity within individual traces.

All of these approaches use token count as a proxy for overthinking---a lossy signal that penalizes all length equally, including legitimate complex reasoning. Our method instead targets \emph{why} a response is long (detecting actual answer drift) rather than \emph{that} it is long.

\subsection{Token- and Segment-Level Credit Assignment}
Standard GRPO broadcasts a single advantage to all tokens. Several methods provide finer-grained credit through learned weighting (GTPO \citep{tan2025gtpo}, $\lambda$-GRPO \citep{wang2025lambdagrpo}), external judges (CAPO \citep{xie2025capo}), or Monte Carlo value estimation (VinePPO \citep{kazemnejad2024vineppo}, SPO \citep{guo2025segment}).

Most closely related to our work, VPPO \citep{liu2026vppo} uses a process reward model (PRM) to localize the first incorrect step in a failed rollout, partitioning the trajectory into a verified correct prefix (rewarded) and an erroneous suffix (penalized). This shares our intuition that not all tokens in a failed trace deserve equal blame. However, VPPO relies on an external PRM for error localization and targets \emph{correctness} boundaries, whereas our method uses the model's own intermediate answers to detect \emph{drift} boundaries---requiring no additional models and applying to both correct and incorrect traces, since a correct trace that drifts before self-correcting still wastes compute. VPPO's binary prefix/suffix partition also does not capture the richer structure of traces with multiple answer changes, which our segment-based formulation handles naturally.

\subsection{Inference-Time Early Exit}
Orthogonal to our training-time approach, several methods diagnose overthinking at inference \citep{fu2024certaindex,yang2025deer,rom2026,muennighoff2025s1}. These are complementary to ours: a model trained with drift-aware shaping could additionally use inference-time early exit for further efficiency gains. We discuss extended related work in Appendix~\ref{app:related_work}.

\section{Qualitative comparison of reasoning traces}
\begin{table*}[!t]
\centering
\small
\renewcommand{\arraystretch}{1.3}
\begin{tabularx}{\textwidth}{p{0.08\textwidth}XX}
\toprule
& \textbf{AIME 2025 Problem \#1} & \textbf{AIME 2025 Problem \#2} \\
& \footnotesize Let $N$ be the number of 8-digit positive integers using digits $1$--$8$ exactly once divisible by 22. Find $N - 2025$. \quad (\textbf{Answer: 279})
& \footnotesize Let $A$ be the set of positive integer divisors of 2025 and $B$ a random subset of $A$. The probability that $B$ is nonempty with $\mathrm{lcm}(B)=2025$ is $m/n$. Find $m+n$. \quad (\textbf{Answer: 237}) \\
\midrule
\raggedright Base Model
&
\raggedright
{\color{correctgreen}$\checkmark$~\textit{``2304 $-$ 2025 = 279.''}} \newline
{\color{driftred}$\circlearrowleft$~\textit{``Is that the answer? Wait, let me verify\ldots''}} \newline
{\color{correctgreen}$\checkmark$~\textit{``Then, 2304 $-$ 2025 = 279.''}} \newline
{\color{driftred}$\circlearrowleft$~\textit{``Is that correct? Let me verify with another approach\ldots''}} \newline
{\color{correctgreen}$\checkmark$~\textit{``Maybe the answer is 279, but I need to verify.''}} \newline
{\color{driftred}$\circlearrowleft$~\textit{``Wait, let's check another resource\ldots''}} \newline
\hfill $\cdots$ \newline
{\color{driftred}\ding{55}~\textbf{Token limit exhausted. No final answer.}}
&
\raggedright
{\color{correctgreen}$\checkmark$~\textit{``m + n = 237.''}} \newline
{\color{driftred}$\circlearrowleft$~\textit{``Wait, hold on. Let me double-check\ldots''}} \newline
{\color{correctgreen}$\checkmark$~\textit{``Therefore, m + n = 237.''}} \newline
{\color{driftred}$\circlearrowleft$~\textit{``But wait, let me think again. Let's think differently\ldots''}} \newline
\hfill $\cdots$ \newline
{\color{driftred}\ding{55}~\textbf{Token limit exhausted. No final answer.}}
\tabularnewline
\midrule
\raggedright \textbf{DASH}
&
\raggedright
{\color{correctgreen}$\checkmark$~\textit{``So 2304 $-$ 2025 = 279.''}} \newline
{\color{gray}$\triangleright$~\textit{``Wait, let me double-check the logic\ldots''}} \newline
{\color{correctgreen}$\checkmark$~\textit{``The calculation seems correct.''}} \newline
{\color{correctgreen}\ding{51}~$\boxed{279}$\textbf{~~Correct.}}
&
\raggedright
{\color{correctgreen}$\checkmark$~\textit{``The answer is 109 + 128 = 237.''}} \newline
{\color{gray}$\triangleright$~\textit{``Let me just make sure I didn't make a mistake\ldots''}} \newline
{\color{correctgreen}$\checkmark$~\textit{``So m + n = 237.''}} \newline
{\color{correctgreen}\ding{51}~$\boxed{237}$\textbf{~~Correct.}}
\tabularnewline
\bottomrule
\end{tabularx}
\caption{Qualitative comparison of reasoning traces on two AIME~2025 problems. The base model reaches the correct answer repeatedly ({\color{correctgreen}$\checkmark$}) but enters self-verification loops ({\color{driftred}$\circlearrowleft$}) until the token limit is exhausted. DASH commits to the answer after brief verification ({\color{gray}$\triangleright$}), avoiding drift entirely.}
\label{tab:drift_case_studies}
\end{table*}